\documentclass[letterpaper]{article} 
\usepackage{aaai2026}  
\usepackage{times}  
\usepackage{helvet}  
\usepackage{courier}  
\usepackage[hyphens]{url}  
\usepackage{graphicx} 
\usepackage{natbib}  
\usepackage{caption} 
\usepackage{algorithm}
\usepackage{algorithmic}

\usepackage{newfloat}
\usepackage{listings}
\DeclareCaptionStyle{ruled}{labelfont=normalfont,labelsep=colon,strut=off} 
\floatstyle{ruled}
\newfloat{listing}{tb}{lst}{}
\floatname{listing}{Listing}
\usepackage[utf8]{inputenc} 
\usepackage[T1]{fontenc}    
\usepackage{hyperref}       
\usepackage{url}            
\usepackage{booktabs}       
\usepackage{amsfonts}       
\usepackage{nicefrac}       
\usepackage{microtype}      
\usepackage{xcolor}         
\usepackage{graphicx}
\usepackage{caption}
\usepackage{subcaption}
\usepackage{tcolorbox}
\usepackage{lineno}
\usepackage{colortbl}

\definecolor{mylightblue}{RGB}{200, 210, 250} 
\definecolor{darkblue}{rgb}{0, 0, 0.5}
\hypersetup{colorlinks=true, citecolor=darkblue, linkcolor=darkblue, urlcolor=darkblue}

\usepackage{amsmath}
\usepackage[capitalize,noabbrev]{cleveref}

\title{Reinforcement Learning for Reasoning in Small LLMs: What Works and What Doesn't}

\author{Quy-Anh Dang$^{1,2}$, Chris Ngo$^{2}$ \\
}  

\affiliations{

    $^1$VNU University of Science, Vietnam \\
    $^2$Knovel Engineering Lab, Singapore \\
    \texttt{\{quyanh.dang, chris.ngo\}@knoveleng.com}
}

\usepackage{bibentry}

\begin{document}

\maketitle

\begin{abstract}
Enhancing the reasoning capabilities of large language models (LLMs) typically relies on massive computational resources and extensive datasets, limiting accessibility for resource-constrained settings. Our study investigates the potential of reinforcement learning (RL) to improve reasoning in small LLMs, focusing on a 1.5-billion-parameter model, \texttt{DeepSeek-R1-Distill-Qwen-1.5B}, under strict constraints: training on 4 NVIDIA A40 GPUs (48 GB VRAM each) within 24 hours. Adapting the Group Relative Policy Optimization (GRPO) algorithm and curating a compact, high-quality mathematical reasoning dataset, we conducted three experiments to explore model behavior and performance. Our results demonstrate rapid reasoning gains - e.g., AMC23 accuracy rising from 63\% to 80\% and AIME24 reaching 46.7\%, surpassing \texttt{o1-preview} - using only 7,000 samples and a \$42 training cost, compared to thousands of dollars for baseline models. However, challenges such as optimization instability and length constraints emerged with prolonged training. These findings highlight the efficacy of RL-based fine-tuning for small LLMs, offering a cost-effective alternative to large-scale approaches. We release our code and datasets as open-source resources, providing insights into trade-offs and laying a foundation for scalable, reasoning-capable LLMs in resource-limited environments. All are available at \textcolor{blue}{\url{https://github.com/knoveleng/open-rs}}.
\end{abstract}

\section{Introduction}
\label{sec:intro}

Recent advancements in large language models (LLMs) have significantly advanced the pursuit of artificial general intelligence (AGI), with models such as GPT-4o~\citep{gpt4o}, Claude 3.5 Sonnet~\citep{claude35sonnet}, and Gemini 1.5~\citep{gemini1_5} demonstrating unprecedented capabilities. A pivotal aspect of this progress is the integration of post-training techniques into the training pipeline. These methods - including supervised fine-tuning (SFT) and reinforcement learning (RL) - enhance reasoning accuracy, align models with societal values, and adapt them to user preferences, all while demanding fewer computational resources than pre-training~\citep{o1}. A notable innovation in this domain is OpenAI’s o1 series, which leverages inference-time scaling through extended Chain-of-Thought (CoT) reasoning to achieve remarkable performance in mathematics, coding, and scientific reasoning tasks~\citep{o1}. However, despite these breakthroughs, scaling reasoning capabilities at test time remains a persistent challenge for the broader research community, largely due to limited access to proprietary methodologies and resources.

Efforts to bolster LLM reasoning have explored diverse strategies. Process-based reward models~\citep{uesato2022solving,lightman2023let,mathshepherd} guide models toward structured problem-solving, while RL approaches~\citep{kumar2024training} optimize performance through feedback-driven learning. Search algorithms, such as Monte Carlo Tree Search (MCTS) and Beam Search, have also been employed to enhance reasoning depth~\citep{feng2024alphazeroliketreesearchguidelarge,xin2024deepseekproverv15harnessingproofassistant,AlphaGeometryTrinh2024}. Although these methods have driven incremental gains, they fall short of the general reasoning prowess exhibited by the o1 series. Recently, the DeepSeek-R1 model~\citep{deepseekr12025} has emerged as a competitive alternative, utilizing RL with the Group Relative Policy Optimization (GRPO) algorithm. Built on the 671-billion-parameter DeepSeek-V3, DeepSeek-R1 matches o1’s reasoning performance~\citep{deepseekr12025}. Yet, the sheer scale and computational demands of such models - often exceeding hundreds of billions of parameters - render them impractical for self-hosting by most organizations outside major technology firms, limiting their broader adoption.

In contrast, small LLMs, typically ranging from 1 to 10 billion parameters, present a resource-efficient alternative with potential for widespread deployment. Previous studies have demonstrated the feasibility of enhancing small LLMs through RL-based fine-tuning inspired by DeepSeek-R1~\citep{deepscaler2025,Slow_Thinking_with_LLMs_3_Preview}. However, these efforts often rely on expansive datasets (hundreds of thousands to millions of samples) or incur significant computational costs, undermining their accessibility for resource-constrained settings. This tension motivates two central research questions:

\begin{enumerate}
    \item \textit{How do small LLMs behave when fine-tuned under strict resource constraints, such as limited computational power and training time?}
    \item \textit{Can their reasoning performance be elevated using an RL-based approach akin to DeepSeek-R1’s methodology, and if so, how?}
\end{enumerate}

These questions naturally extend to a practical inquiry: If viable, how should such an approach be implemented for small LLMs, and if not, what are the fundamental limitations? Addressing these, we investigate the reasoning capacity of a 1.5-billion-parameter model, \texttt{DeepSeek-R1-Distill-Qwen-1.5B}, under stringent constraints: training on a cluster of 4 NVIDIA A40 GPUs (48 GB VRAM each) within a 24-hour window. Our methodology adapts the GRPO-based RL framework from DeepSeek-R1, tailoring it to the resource-limited context of small LLMs. We assess performance on a suite of mathematical reasoning benchmarks, a domain requiring structured, logical problem-solving that serves as a robust testbed for reasoning ability.

\begin{figure}[ht]
\centering
\includegraphics[width=0.48\textwidth]{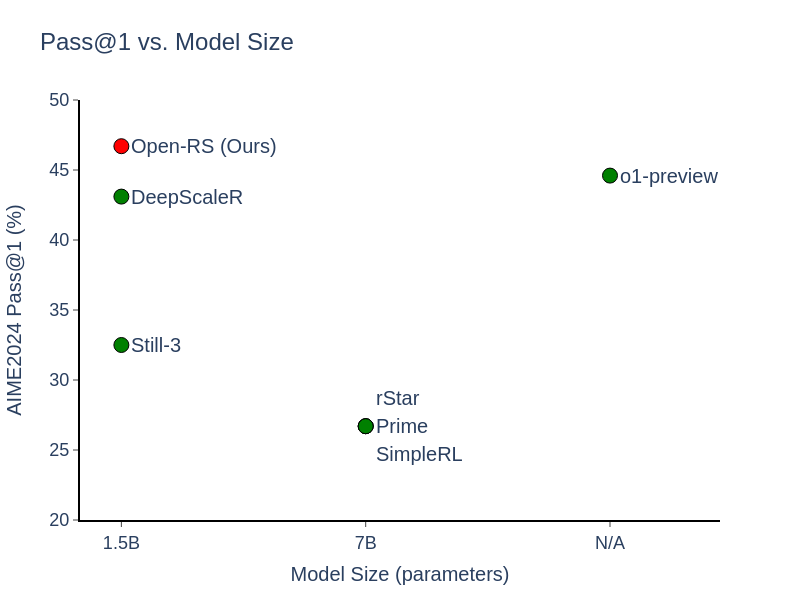}
\includegraphics[width=0.48\textwidth]{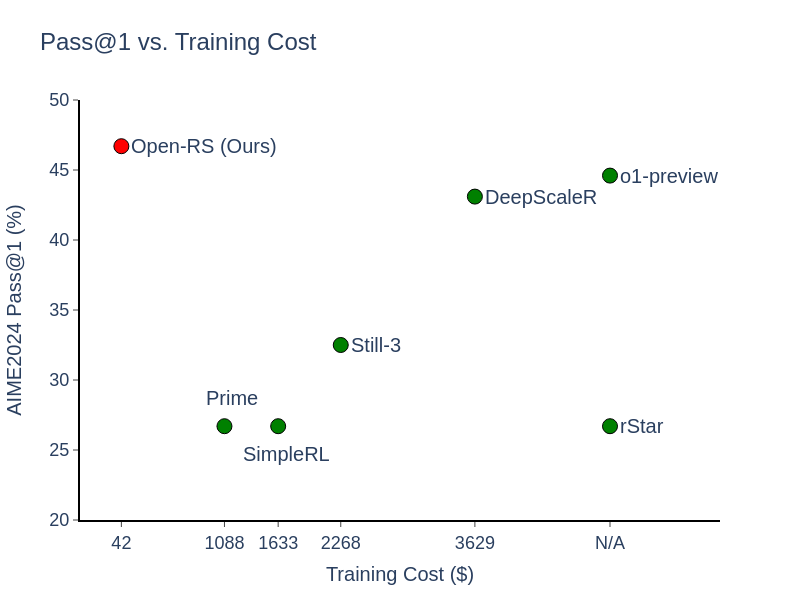}
\caption{Comparison of zero-shot pass@1 performance versus model size (left) and computational cost (right). Our Open-RS (\textcolor{red}{red point}) achieves the highest AIME24 score (46.7\%), outperforming \texttt{o1-preview} (44.6\%) and other models (\textcolor{green}{green points}). Additionally, Open-RS models exhibit the lowest computational cost at approximately \$42.}
\label{fig:overall-diagram}
\end{figure}

Our study yields three primary contributions:
\begin{enumerate}
    \item We systematically analyze the reasoning potential of small LLMs under specific computational constraints, providing a practical lens on their scalability and deployment feasibility.
    \item We offer actionable insights into the efficacy and challenges of RL-based fine-tuning for small LLMs, bridging the gap between theoretical advancements and real-world applicability.
    \item We release our source code and curated datasets as open-source resources, fostering reproducibility and encouraging further exploration by the research community.
\end{enumerate}

Our findings illuminate the promise of RL-based methods to enhance small LLMs’ reasoning capabilities, achieving competitive performance with minimal resources (\cref{fig:overall-diagram}). Simultaneously, they reveal critical challenges - such as data efficiency, optimization stability, and length constraints - that must be addressed to fully realize this potential. These insights lay the groundwork for developing lightweight, reasoning-capable LLMs suitable for resource-constrained environments, advancing the democratization of advanced AI technologies.

The remainder of this paper is structured as follows: Section~\ref{sec:methodology} details our methodology, including data curation, RL algorithm, and reward design; Section~\ref{sec:experiments} presents three experiments, their results, and comparative analyses; and Section~\ref{sec:conclusion} summarizes key findings. Additional details, including related work, discussion, hyperparameter setups, and supplementary results, are provided in the Appendix.


\section{Methodology}
\label{sec:methodology}

In this section, we outline our approach to optimizing the reasoning capabilities of small large language models (LLMs) under computational constraints. Our methodology comprises two primary components: (1) the curation of a high-quality, mathematics-focused dataset, and (2) the application of a resource-efficient reinforcement learning (RL) algorithm. These components are designed to balance performance gains with practical limitations, such as reduced computational overhead and privacy considerations.

\subsection{High-Quality Dataset Curation}
\label{subsec:dataset_curation}

To minimize training costs while maximizing reasoning performance, we curate a compact, high-quality dataset tailored to mathematical reasoning. This dataset is derived from two existing sources: the s1 dataset~\citep{s1simpletesttimescaling} and the DeepScaleR dataset~\citep{deepseekr12025}. By filtering and refining these datasets, we ensure that our training data is both relevant and challenging, enabling efficient learning for small LLMs.

\paragraph{s1 Dataset}
\label{subsubsec:s1_dataset}
The s1 dataset~\citep{s1simpletesttimescaling} is a general-purpose reasoning corpus comprising 59,029 questions sourced from diverse domains, including NuminaMATH~\citep{numina_math_datasets}, AIME problems (1983--2021), OlympicArena~\citep{huang2024olympicarena}, OmniMath~\citep{gao2024omnimath}, AGIEval~\citep{zhong2023agieval}, probability questions from Stanford University's Statistics Department PhD Qualifying Exams (\url{https://statistics.stanford.edu}), and brain-teasers from PuzzledQuant (\url{https://www.puzzledquant.com}). Although the dataset spans multiple disciplines - such as Astronomy, Biology, Chemistry, Computer Science, Geography, Mathematics, and Physics - our focus is exclusively on mathematical reasoning. 

To isolate mathematics-specific examples, we adopt a filtering workflow inspired by~\citep{s1simpletesttimescaling}. First, we retain only questions with solutions containing the LaTeX command \texttt{\textbackslash boxed\{\}}, a common indicator of mathematical answers, reducing the dataset to 31,323 examples. Next, we employ the distilled model \texttt{DeepSeek-R1-Distill-Qwen-1.5B} to eliminate trivial questions, yielding 21,533 examples. Finally, to ensure data quality, we use \texttt{Qwen2.5-7B-Instruct} to remove noisy or multi-part questions, resulting in a final set of 18,615 high-quality mathematical reasoning examples -- \textbf{open-s1} dataset.

\paragraph{DeepScaleR Dataset}
\label{subsubsec:deepscaler_dataset}
The DeepScaleR dataset~\citep{deepscaler2025} contains 40,315 mathematics-specific questions drawn from AIME (1984--2023), AMC (prior to 2023), Omni-MATH, and the Still dataset. Unlike the s1 dataset, DeepScaleR is pre-filtered to focus solely on mathematics, with redundant questions removed and solutions extracted from raw text using retrieval-augmented generation (RAG) and advanced LLMs like \texttt{Gemini-1.5-Pro-002}. To further refine this dataset, we apply \texttt{Qwen2.5-Math-7B-Instruct} to exclude easy questions, reducing the set to 21,044 examples  -- \textbf{open-deepscaler} dataset. We opt for \texttt{Qwen2.5-Math-7B-Instruct} over \texttt{DeepSeek-R1-Distill-Qwen-1.5B} - used for the s1 dataset - to introduce diversity in filtering criteria and avoid excessive overlap between the two datasets. 
\paragraph{Final Dataset}
Combining the refined \textbf{open-s1} dataset (18,615 examples) and \textbf{open-deepscaler} (21,044 examples), we obtain a final high-quality dataset of 39,659 mathematical reasoning questions. This curated corpus strikes a balance between scale and specificity, enabling effective training of small LLMs under resource constraints.

\subsection{Reinforcement Learning Algorithm}
\label{subsec:rl_algorithm}

To train small LLMs efficiently, we adopt the Group Relative Policy Optimization (GRPO) algorithm~\cite{deepseekmath}, as utilized in~\cite{deepseekr12025}. GRPO eliminates the need for a separate critic model - typically as large as the policy model - by estimating baselines from group scores, thereby reducing computational overhead. For each question \( q \), GRPO samples a group of \( G \) outputs \( \{o_1, o_2, \dots, o_G\} \) from the old policy \( \pi_{\theta_{\text{old}}} \) and optimizes the policy \( \pi_{\theta} \) by maximizing the following objective:

\begin{align}
    \mathcal{J}_{\text{GRPO}}(\theta) &= \mathbb{E}_{[q \sim P(Q), \{o_i\}_{i=1}^G \sim \pi_{\theta_{\text{old}}}(O|q)]} \nonumber \\
    & \quad \frac{1}{G} \sum_{i=1}^G \left( \min \left( \frac{\pi_\theta(o_i | q)}{\pi_{\theta_{\text{old}}}(o_i | q)} A_i, \right. \right. \nonumber \\
    & \quad \quad \left. \left. \text{clip} \left( \frac{\pi_\theta(o_i | q)}{\pi_{\theta_{\text{old}}}(o_i | q)}, 1 - \epsilon, 1 + \epsilon \right) A_i \right) \right) \nonumber \\
    & \quad - \beta \mathbb{D}_{\text{KL}}(\pi_{\theta} || \pi_{\text{ref}}) \label{eq:grpo_obj}
\end{align}
where the KL-divergence term is defined as:
\begin{equation}
    \mathbb{D}_{\text{KL}}(\pi_{\theta} || \pi_{\text{ref}}) = \frac{\pi_{\text{ref}}(o_i|q)}{\pi_{\theta}(o_i|q)} - \log \frac{\pi_{\text{ref}}(o_i|q)}{\pi_{\theta}(o_i|q)} - 1,
\end{equation}
and the advantage \( A_i \) is computed from a group of rewards \( \{r_1, r_2, \dots, r_G\} \):
\begin{equation}
    A_i = \frac{r_i - \text{mean}(\{r_1, r_2, \dots, r_G\})}{\text{std}(\{r_1, r_2, \dots, r_G\})}.
\end{equation}
Here, \( \epsilon \) and \( \beta \) are hyperparameters controlling the clipping range and KL penalty, respectively.

\paragraph{Reward Models}
\label{subsubsec:reward_models}
The reward function is critical to guiding RL optimization. We employ a rule-based reward system comprising three components, designed to balance correctness, efficiency, and structure without relying on resource-intensive neural reward models:

\begin{itemize}
    \item \textbf{Accuracy Reward}: This evaluates whether the model’s response is correct, requiring the final answer to be presented in a \texttt{\textbackslash boxed\{\}} format for reliable verification. A binary score (1 for correct, 0 for incorrect) ensures simplicity and objectivity.
    \item \textbf{Cosine Reward}: This augments the accuracy reward by scaling it based on response length using a cosine schedule. Shorter correct solutions receive higher rewards, while longer incorrect solutions are penalized less severely, incentivizing concise yet accurate reasoning.
    \item \textbf{Format Reward}: This enforces structural clarity by requiring the model to encapsulate its reasoning process within \texttt{<think>} and \texttt{</think>} tags, awarding a positive score for compliance.
\end{itemize}


\section{Experiments}
\label{sec:experiments}

To address the research questions outlined in Section~\ref{sec:intro} - namely, how reinforcement learning (RL) can enhance the reasoning abilities of small large language models (LLMs) and what practical insights emerge under computational constraints - we design three experiments to analyze the training behavior of small LLMs. These experiments aim to provide empirical evidence of performance improvements and offer actionable guidance for future research and industrial applications.

\subsection{Experimental Setup}
\label{subsec:setup}

We select \texttt{DeepSeek-R1-Distill-Qwen-1.5B}~\citep{deepseekr12025} as our base model for training. This 1.5-billion-parameter model, distilled from larger architectures, is chosen for its balance of efficiency and reasoning potential. Notably, we bypass the supervised fine-tuning (SFT) phase - typically a precursor to RL for performance enhancement~\citep{chu2025sftmemorizesrlgeneralizes} - hypothesizing that the model’s pretraining is sufficient to leverage RL directly. For the RL phase, we employ the Group Relative Policy Optimization (GRPO) algorithm, as detailed in Section~\ref{subsec:rl_algorithm}, due to its computational efficiency.

Training is conducted on a cluster of 4 NVIDIA A40 GPUs (48GB VRAM each), imposing constraints that limit us to sampling 6 outputs per step with a maximum completion length of 4096 tokens. To facilitate this, we adapt \texttt{open-r1}~\citep{openr1}, an open-source reproduction of DeepSeek-R1 by the Hugging Face team, customizing it to align with our objectives. The training phase is restricted to 1 epoch, completed within a 24-hour window, reflecting real-world resource limitations. Hyperparameters and additional configurations are detailed in Appendix~\ref{app:hyperparameters}.

\subsection{Benchmark Datasets}
\label{subsec:benchmarks}
To evaluate the reasoning capabilities of our small LLM, we choose five mathematics-focused benchmark datasets: \textbf{AIME24}~\footnote{\url{https://huggingface.co/datasets/AI-MO/aimo-validation-aime}}, \textbf{MATH-500}~\citep{lightman2023lets,hendrycksmath2021}, \textbf{AMC23}~\footnote{\url{https://huggingface.co/datasets/AI-MO/aimo-validation-amc}}, \textbf{Minerva}~\citep{minervamath} and \textbf{OlympiadBench}~\citep{he-etal-2024-olympiadbench}. Details of the datasets are provided in \cref{app:benchmarks}.

\subsection{Baseline Models}
\label{subsec:baselines}

To contextualize our results, we compare our trained model against a range of baselines: \texttt{Llama-3.1-70B-Instruct}~\citep{MetaAI2024}, \texttt{o1-preview}~\citep{o1preview}, \texttt{Qwen-2.5-Math-7B-Instruct}~\citep{yang2024qwen25mathtechnicalreportmathematical}, \texttt{rStar-Math-7B}~\citep{guan2025rstar}, \texttt{Eurus-2-7B-PRIME}, \texttt{Qwen2.5-7B-SimpleRL}~\citep{zeng2025simplerl}~\citep{cui2025processreinforcementimplicitrewards}, \texttt{DeepSeek-R1-Distill-Qwen-1.5B}~\citep{deepseekr12025}, \texttt{DeepScaleR-1.5B-Preview}~\citep{deepscaler2025}, \texttt{Still-3-1.5B-Preview}~\citep{Slow_Thinking_with_LLMs_3_Preview}.

This selection enables a robust comparison across model sizes, training methodologies, and reasoning strategies, highlighting the efficacy of our approach for small LLMs. Details of the baselines are provided in \cref{app:baseline-models}.

\subsection{Evaluation Metric}
\label{subsec:metric}

We adopt the zero-shot pass@1 metric to measure performance, defined as the proportion of problems correctly solved on the first attempt without prior examples. This metric emphasizes the model’s ability to reason independently, aligning with our goal of enhancing intrinsic reasoning capabilities in small LLMs. Final answers are required in \texttt{\textbackslash boxed\{\}} format for consistent automated evaluation.

\subsection{Process and Results}
\label{subsec:results}

In this subsection, we present three experiments designed to enhance the reasoning abilities of small LLMs using reinforcement learning (RL), follow the methodology in Section~\ref{sec:methodology}. We analyze training progress, evaluate performance across benchmarks, and compare our models against baselines, highlighting key insights and their implications for future work.

\begin{figure}[ht]
\centering
\includegraphics[width=0.45\textwidth]{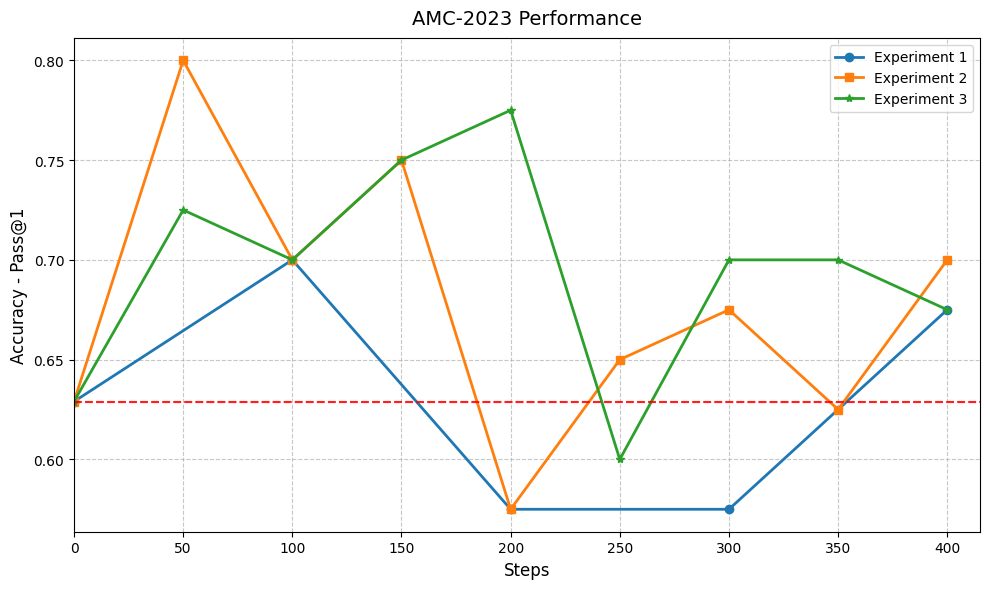}
\includegraphics[width=0.45\textwidth]{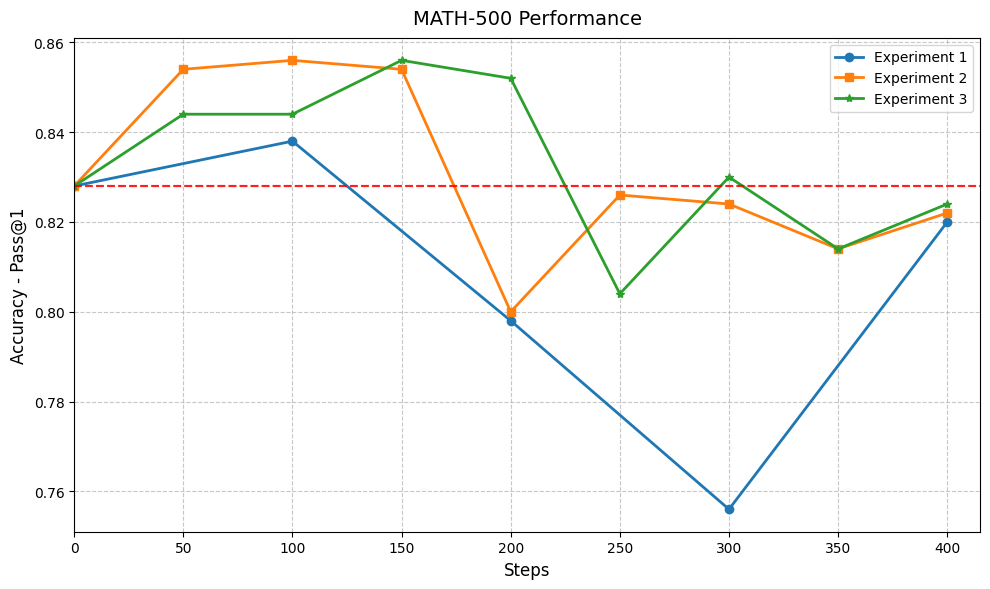}
\caption{Performance of the model on \textbf{AMC23} (left) and \textbf{MATH-500} (right) across global training steps. The \textcolor{red}{red dashed line} indicates the baseline score at the start of training.}
\label{fig:performance-steps}
\end{figure}

\paragraph{Experiment 1: Impact of High-Quality Data.}
In Experiment 1, we train the \texttt{DeepSeek-R1-Distill-Qwen-1.5B} model using the \textbf{open-s1} dataset (18,615 samples) from Section~\ref{subsec:dataset_curation}, with a maximum completion length of 4096 tokens. We employ accuracy and format rewards, as described in Section~\ref{subsec:rl_algorithm}. Although the full dataset corresponds to approximately 1500 global steps for one epoch, computational constraints (24-hour limit on 4x A40 GPUs) restrict training to 500 global steps.

\begin{figure}[ht]
\centering
\includegraphics[width=0.45\textwidth]{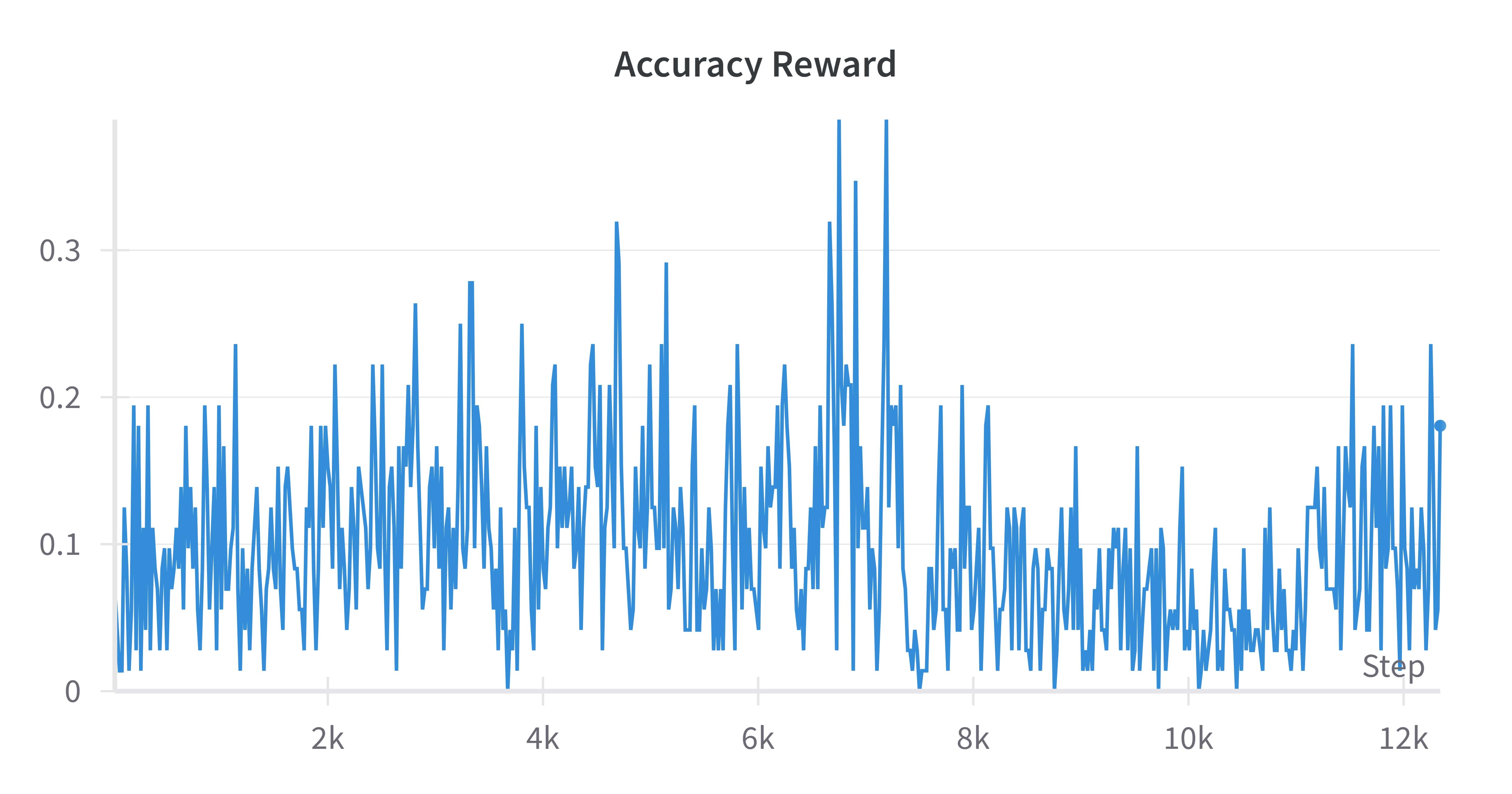}
\includegraphics[width=0.45\textwidth]{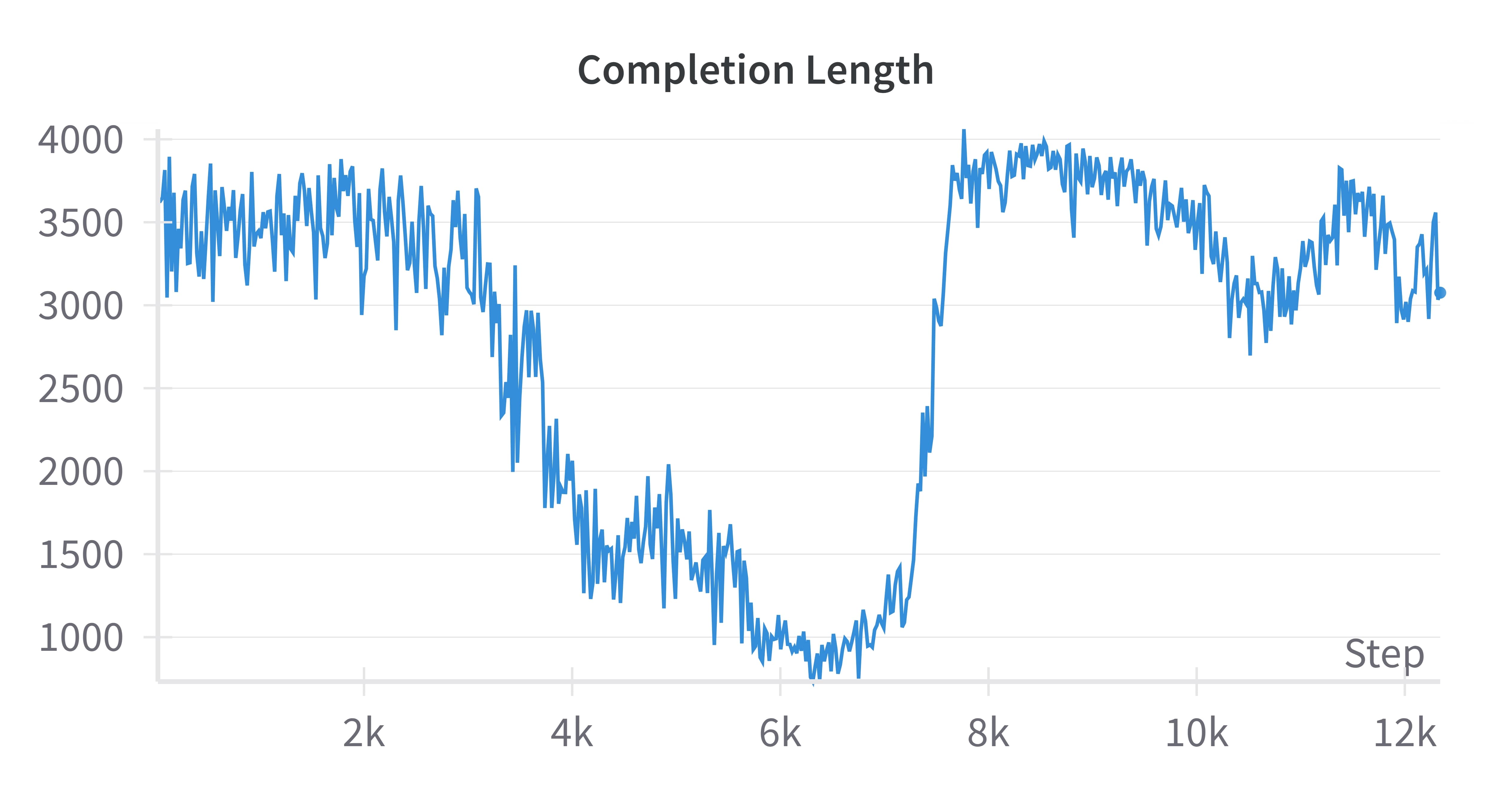}
\caption{Accuracy reward (left) and completion length (right) of outputs in Experiment 1 across local steps. Note that global steps are distributed across 4 GPUs, with 100 global steps approximating 3000 local steps.}
\label{fig:exp1}
\end{figure}

Performance on AMC23 improves from 63\% to 70\% and on MATH-500 from 83\% to 84\% within the first 50–100 steps (see Figure~\ref{fig:performance-steps}). However, after 200 steps, accuracy degrades significantly, dropping below 60\% on AMC23 and to 80\% on MATH-500. Figure~\ref{fig:exp1} illustrates this trend, showing unstable accuracy rewards and completion lengths fluctuating near 4000 tokens initially, then decreasing to around 3000 tokens by 100 global steps (approximately 3000 local steps on a single GPU). Post-200 steps, lengths increase again, accompanied by unreadable content and non-English outputs.

This degradation suggests that the model struggles with the complexity of \textbf{open-s1}, often exceeding the 4096-token limit before producing a final answer. The initial length reduction reflects adaptation to the format reward, but the subsequent increase and language drift indicate reward misalignment. We derive the following insight:

\begin{tcolorbox}[title=Insight 1]
Small LLMs can achieve rapid reasoning improvements with limited high-quality data within 50–100 steps, but performance degrades with prolonged training under strict length constraints.
\end{tcolorbox}

\paragraph{Experiment 2: Balancing Easy and Hard Problems.}
Building on Experiment 1, we hypothesize that mixing easier problems with challenging ones could stabilize training and reduce completion lengths. We construct a dataset of 7000 samples: 3000 from \textbf{open-s1}, 3000 from \textbf{open-deepscaler}, and 1000 easier problems from the raw \textbf{DeepScaleR} dataset (Section~\ref{subsec:dataset_curation}). The maximum completion length is reduced to 3584 tokens, retaining accuracy and format rewards.

\begin{figure}[!htbp]
\centering
\includegraphics[width=0.45\textwidth]{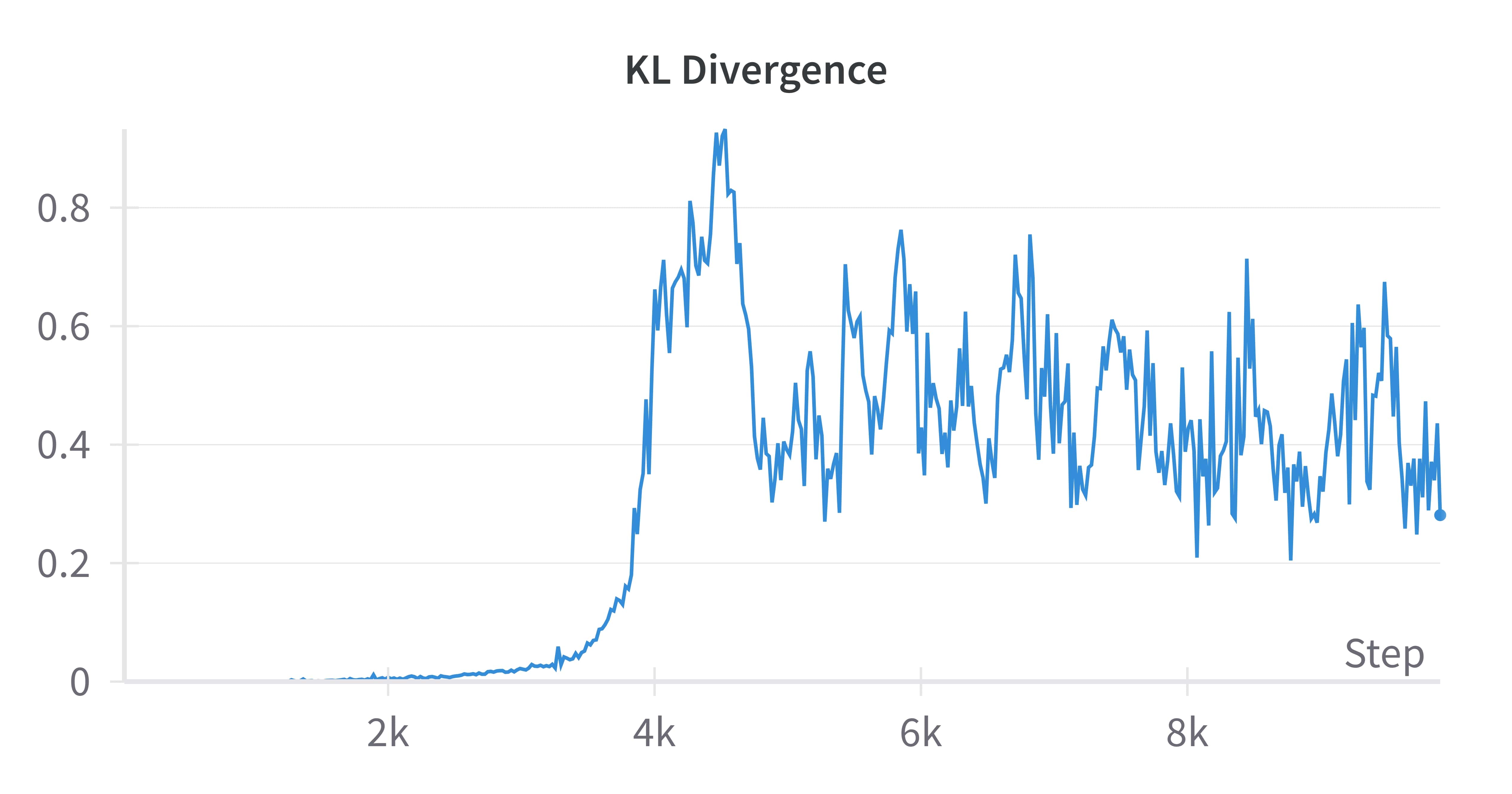}
\includegraphics[width=0.45\textwidth]{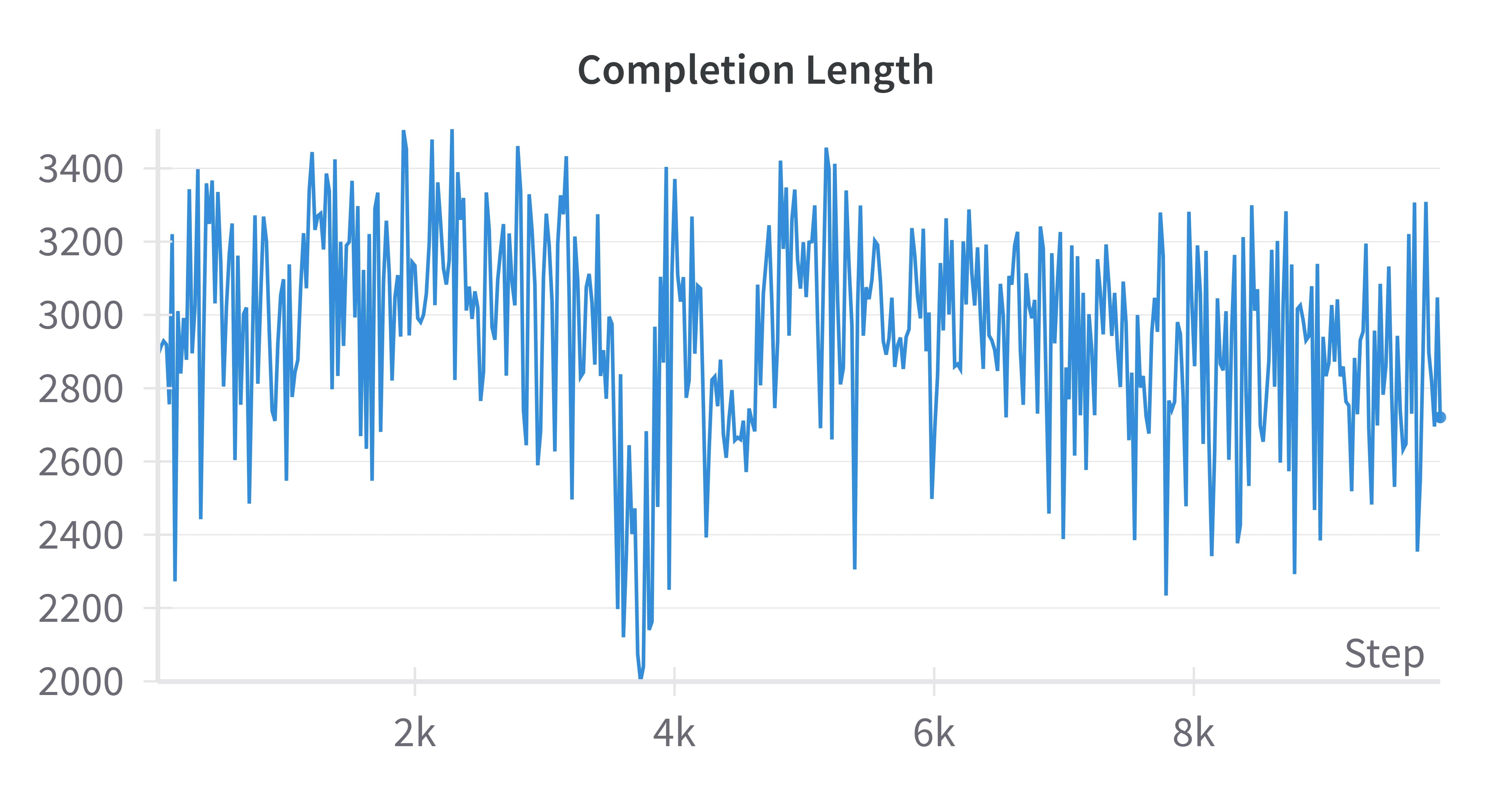}
\caption{KL divergence (left) and completion length (right) of outputs in Experiment 2 across local steps.}
\label{fig:exp2}
\end{figure}

Initial completion lengths drop to approximately 2800 tokens, and performance improves significantly: AMC23 rises from 63\% to 80\%, and MATH-500 from 83\% to 85\% within 50–100 steps (Figure~\ref{fig:performance-steps}). However, after 150–200 steps (approximately 4000 local steps), performance declines, and KL divergence becomes unstable (Figure~\ref{fig:exp2}), with mixed-language outputs reemerging.

The improved initial performance validates our hypothesis, suggesting that easier problems encourage concise reasoning, while harder ones maintain complexity. However, the late-stage instability highlights persistent challenges with length constraints and multilingual tendencies. We note:

\begin{tcolorbox}[title=Insight 2]
Incorporating a mix of easy and hard problems under reduced length constraints enhances early performance and stabilizes reasoning behavior, though long-term stability remains elusive.
\end{tcolorbox}

\paragraph{Experiment 3: Controlling Length with Cosine Reward.}
Experiment 3 uses the same 7000-sample dataset as Experiment 2, but replaces the accuracy reward with a cosine reward to better control output length, as outlined in Section~\ref{subsubsec:reward_models}. We also append an instruction to the system prompt: \textit{“Reply in English only, do not use other languages”}, avoiding a computationally expensive language reward function. The maximum completion length remains 3584 tokens.

\begin{figure}[!htbp]
\centering
\includegraphics[width=0.45\textwidth]{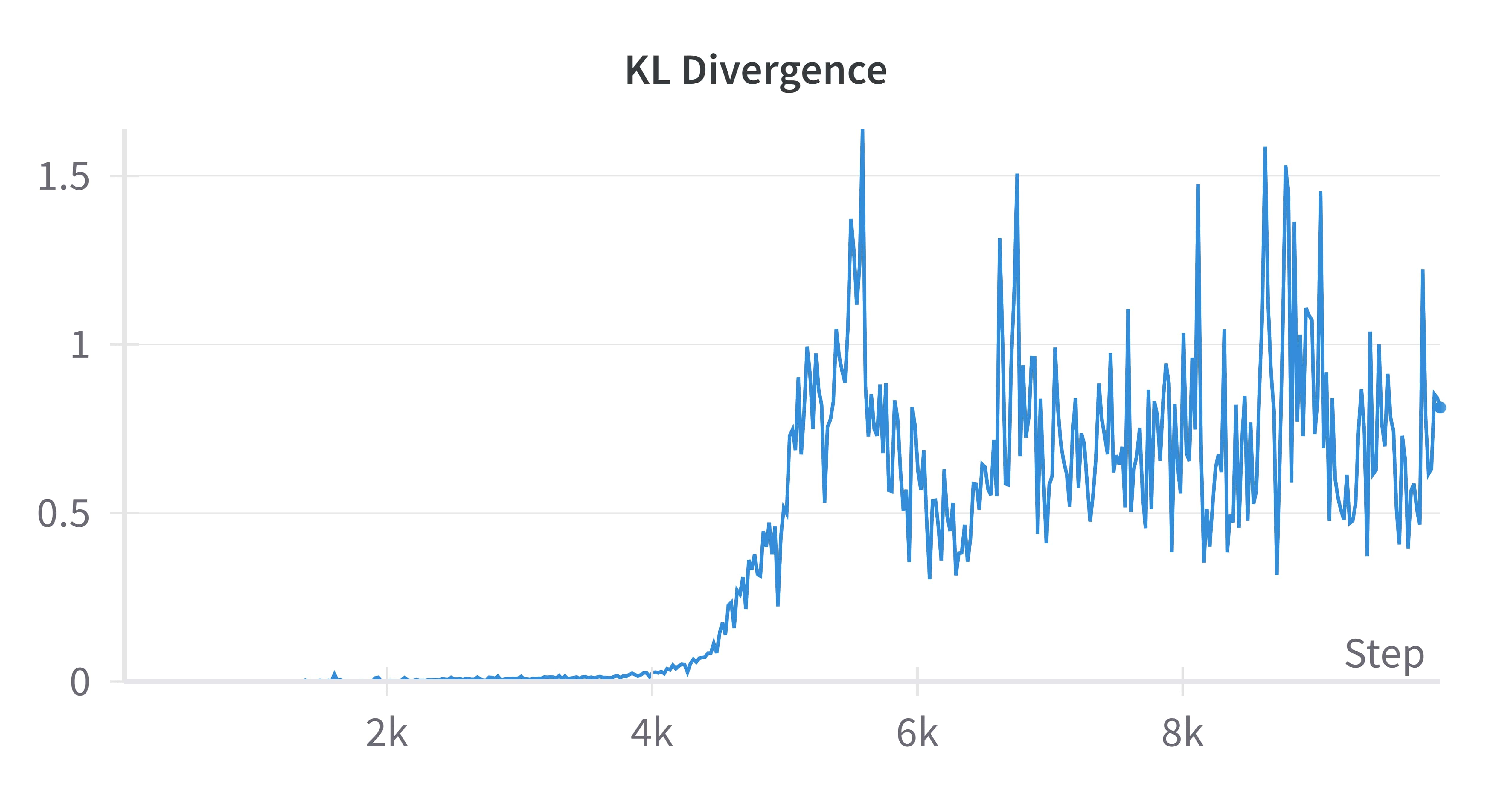}
\includegraphics[width=0.45\textwidth]{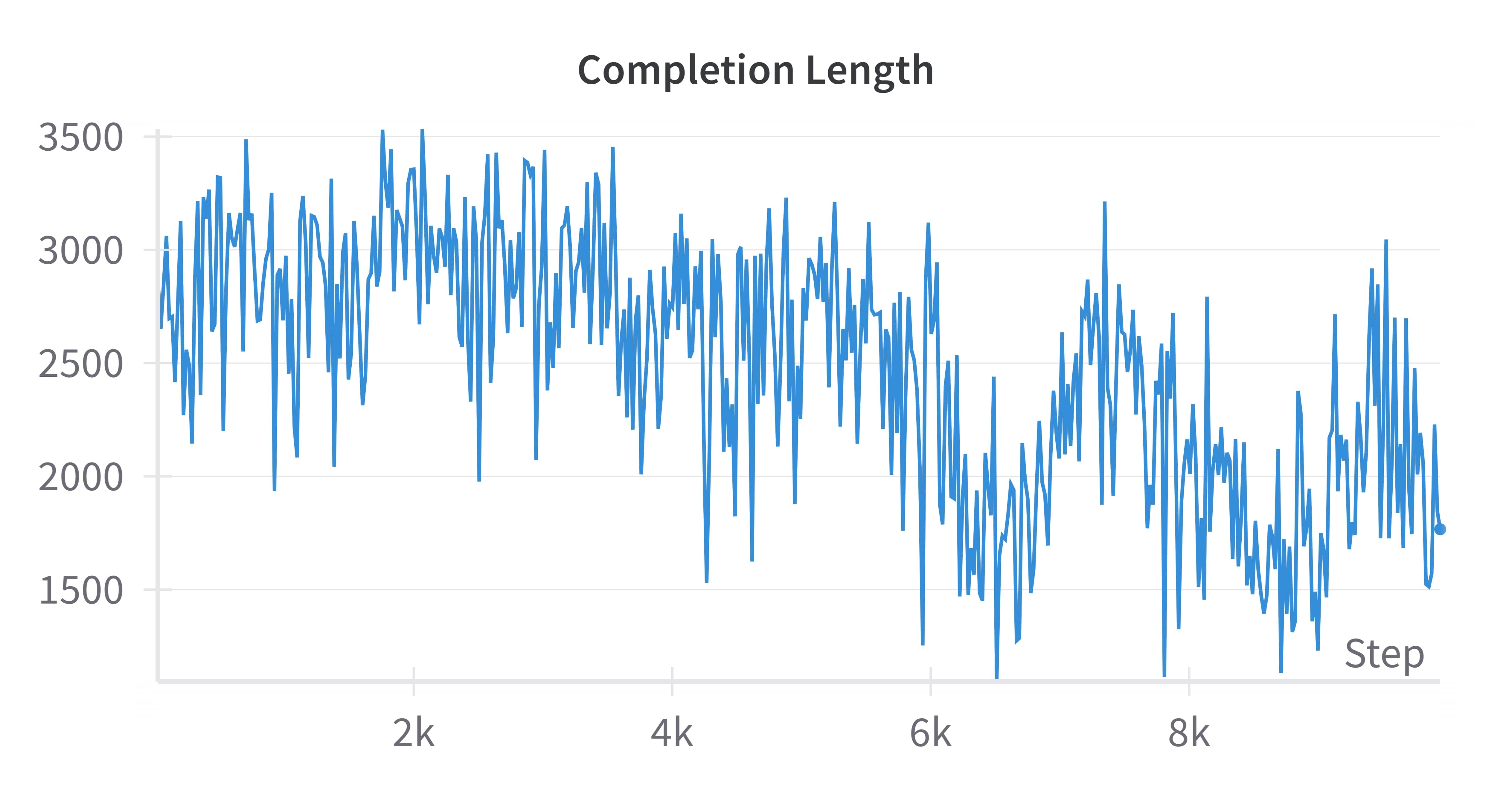}
\caption{KL divergence (left) and completion length (right) of outputs in Experiment 3 across local steps.}
\label{fig:exp3}
\end{figure}

Completion lengths stabilize between 1000 and 3500 tokens (Figure~\ref{fig:exp3}), a marked improvement over Experiment 2’s 2000–3500 range. Performance on AMC23 and MATH-500 increases modestly compared to the baseline (63\% to 72.5\% and 83\% to 84.4\%, respectively) within 50 steps, though it lags behind Experiment 2’s peak (Figure~\ref{fig:performance-steps}). After 200 steps, mixed-language content persists, reflecting the multilingual nature of \texttt{DeepSeek-R1-Distill-Qwen-1.5B}.

The cosine reward effectively regulates length, but the language issue suggests a need for explicit language constraints or extended completion lengths for complex tasks. We conclude:

\begin{tcolorbox}[title=Insight 3]
Cosine rewards stabilize completion lengths, improving training consistency, but extending length limits is necessary for extremely hard tasks, particularly with multilingual base models.
\end{tcolorbox}

\paragraph{Overall Comparison.}
We select checkpoints at 100, 50, and 50 global steps from Experiments 1, 2, and 3, naming them \textbf{Open-RS1}, \textbf{Open-RS2}, and \textbf{Open-RS3} (R for \textbf{Reasoning}, S for \textbf{Small}), respectively. These are evaluated against baselines from Section~\ref{subsec:baselines} across benchmarks from Section~\ref{subsec:benchmarks}, using zero-shot pass@1 (Table~\ref{tab:model_performance}).

\begin{table*}[ht]
\centering
\setlength{\tabcolsep}{2pt}
\renewcommand{\arraystretch}{1.1}
\small
\begin{tabular}{lcccccc}
\toprule
\textbf{Model} & \textbf{AIME24} & \textbf{MATH-500} & \textbf{AMC23} & \textbf{Minerva} & \textbf{OlympiadBench} & \textbf{Avg.} \\
\midrule
\textit{General Models} \\
\midrule
Llama-3.1-70B-Instruct & 16.7 & 64.6 & 30.1 & 35.3 & 31.9 & 35.7 \\
o1-preview & 44.6 & 85.5 & -- & -- & -- & -- \\
\midrule
\textit{7B Models} \\
\midrule
Qwen-2.5-Math-7B-Instruct & 13.3 & 79.8 & 50.6 & 34.6 & 40.7 & 43.8 \\
rStar-Math-7B & 26.7 & 78.4 & 47.5 & -- & 47.1 & -- \\
Eurus-2-7B-PRIME & 26.7 & 79.2 & 57.8 & 38.6 & 42.1 & 48.9 \\
Qwen2.5-7B-SimpleRL & 26.7 & 82.4 & 62.5 & \textbf{39.7} & 43.3 & 50.9 \\
\midrule
\textit{1.5B Models} \\
\midrule
DeepSeek-R1-Distill-Qwen-1.5B & 28.8 & 82.8 & 62.9 & 26.5 & 43.3 & 48.9 \\
Still-3-1.5B-Preview & 32.5 & 84.4 & 66.7 & 29.0 & 45.4 & 51.6 \\
DeepScaleR-1.5B-Preview & 43.1 & \textbf{87.8} & 73.6 & 30.2 & 50.0 & \textbf{57.0} \\
\midrule
\rowcolor{mylightblue!40}
\textit{Our Models} \\
\midrule
\rowcolor{mylightblue!40}
Open-RS1 (100 steps) & 30.0 & 83.8 & 70.0 & 29.0 & \textbf{52.4} & 53.0 \\
\rowcolor{mylightblue!40}
Open-RS2 (50 steps) & 30.0 & 85.4 & \textbf{80.0} & 30.5 & \textbf{52.4} & 55.7 \\
\rowcolor{mylightblue!40}
Open-RS3 (50 steps) & \textbf{46.7} & 84.4 & 72.5 & 26.8 & 51.3 & 56.3 \\
\bottomrule
\end{tabular}
\caption{Zero-shot pass@1 performance across benchmarks. \textbf{Bold} indicates the highest score per benchmark. Dashes (--) denote unavailable official scores. Scores for \texttt{o1-preview} are sourced from~\cite{o1preview}; others from~\cite{zeng2025simplerl,deepscaler2025}. Our models are evaluated using the \texttt{lighteval} package~\cite{lighteval}.}
\label{tab:model_performance}
\end{table*}

Our models outperform most baselines, with average scores of 53.0\% (Open-RS1), 55.7\% (Open-RS2), and 56.3\% (Open-RS3), compared to 57.0\% for \texttt{DeepScaleR-1.5B-Preview}. Notably, Open-RS3 achieves the highest AIME24 score (46.7\%), surpassing \texttt{o1-preview} (44.6\%) and \texttt{DeepScaleR-1.5B-Preview} (43.1\%). Open-RS2 excels on AMC23 (80.0\%) and ties with Open-RS1 on OlympiadBench (52.4\%), both outperforming \texttt{DeepScaleR-1.5B-Preview}. MATH-500 scores remain competitive, though Minerva performance lags behind 7B models, reflecting the complexity of cross-disciplinary reasoning.

\begin{table*}[ht]
\centering
\setlength{\tabcolsep}{2pt}
\renewcommand{\arraystretch}{1.1}
\small
\begin{tabular}{|l | p{2.7cm} | p{2.7cm} | p{2.7cm} | p{2.7cm}|}
\toprule
 & \textbf{rStar-Math-7B} & \textbf{Eurus-2-7B-PRIME} & \textbf{Qwen2.5-7B-SimpleRL} & \textbf{Open-RS} \\
\midrule
Base Model & Qwen2.5-Math-7B & Qwen2.5-Math-7B & Qwen2.5-Math-7B & DeepSeek-R1-Distill-Qwen-1.5B \\
SFT Data & 7.3M & 230k & 0 & 0 \\
RM Data & 7k & 0 & 0 & 0 \\
RM & None & Eurus-2-7B-SFT & None & None \\
RL Data & 3.647M $\times$ 16 & 150k $\times$ 4 & 8k $\times$ 8 & 7k $\times$ 6 \\
Hardware & 10x 8 H100 80GB, 15x 4 A100 40GB & 1x 8 A100 80GB & 4x 6 A100 80GB & 1x 4 A40 48GB \\
Time & -- & 72h & 36h & 24h \\
\midrule
\textbf{Cost Est.} & -- & \$1088 & \$1633 & \$42 \\
\bottomrule
\end{tabular}
\caption{Comparison of data usage and training costs for 7B models. Data are sourced from original papers or GitHub issues addressing author's resource constraints.}
\label{tab:7b-models}
\end{table*}

We further compare training costs\footnote{Cost estimates are calculated based on pricing from \url{https://www.runpod.io/pricing}.} and data efficiency (Tables~\ref{tab:7b-models} and~\ref{tab:1.5b-models}, and \cref{fig:overall-diagram}). Our approach, using 7000 samples with 6 outputs per step (42,000 total samples), costs approximately \$42 on 4x A40 GPUs over 24 hours. In contrast, 7B models like \texttt{Qwen2.5-7B-SimpleRL} (\$1633) and \texttt{Eurus-2-7B-PRIME} (\$1088) and 1.5B models like \texttt{DeepScaleR-1.5B-Preview} (\$3629) and \texttt{Still-3-1.5B-Preview} (\$2268) require significantly more resources and data (e.g., 40k × 16 samples for \texttt{DeepScaleR}).

\begin{table*}[ht]
\centering
\setlength{\tabcolsep}{2pt}
\renewcommand{\arraystretch}{1.1}
\small
\begin{tabular}{|l | p{3.5cm} | p{3.5cm} | p{3.5cm}|}
\toprule
 & \textbf{DeepScaleR-1.5B-Preview} & \textbf{Still-3-1.5B-Preview} & \textbf{Open-RS} \\
\midrule
Base Model & DeepSeek-R1-Distill-Qwen-1.5B & DeepSeek-R1-Distill-Qwen-1.5B & DeepSeek-R1-Distill-Qwen-1.5B \\
SFT Data & 0 & 0 & 0 \\
RM Data & 0 & 0 & 0 \\
RM & None & None & None \\
RL Data & 40k $\times$ 16 & 30k $\times$ 8 & 7k $\times$ 6 \\
Hardware & 8x A100 80GB & 1x 8 A100 80GB & 1x 4 A40 48GB \\
Time & 240h & 150h & 24h \\
\midrule
\textbf{Cost Est.} & \$3629 & \$2268 & \$42 \\
\bottomrule
\end{tabular}
\caption{Comparison of data usage and training costs for 1.5B models. Data are sourced from original papers or GitHub issues addressing author's resource constraints.}
\label{tab:1.5b-models}
\end{table*}

Our approach demonstrates that small LLMs can achieve competitive reasoning performance with minimal data and cost, offering a scalable alternative to resource-intensive baselines.

\section{Conclusion}
\label{sec:conclusion}

Our study investigated enhancing the reasoning abilities of small LLMs using RL, focusing on the 1.5-billion-parameter \texttt{DeepSeek-R1-Distill-Qwen-1.5B} under strict constraints. Adapting the GRPO algorithm and a compact mathematical reasoning dataset, we conducted three experiments to assess behavior and performance under resource limitations. Our findings show small LLMs can achieve significant reasoning gains with minimal resources - e.g., AMC23 accuracy rising from 63\% to 80\% and AIME24 reaching 46.7\%, surpassing \texttt{o1-preview} - at a cost of \$42 versus thousands for baselines. Open-RS variants averaged 53.0\%–56.3\% on benchmarks, demonstrating RL’s viability for small LLMs. Releasing our code and datasets, we provide a framework for lightweight, reasoning-capable models, despite challenges like optimization stability, laying a foundation for future work.

\bibliography{main}


\newpage
\tableofcontents
\newpage
\appendix \label{app:appendix}
\section{Related Work}
\label{sec:related_work}

\subsection{Reasoning in Large Language Models}
\label{subsec:reasoning_llms}
A substantial body of research has investigated methods to enhance the reasoning capabilities and factual accuracy of large language models (LLMs). Early approaches predominantly relied on prompting techniques to elicit structured reasoning. For instance, scratchpad-style prompting encourages models to break down problems into intermediate steps~\citep{nye2021show}, while verification mechanisms assess the correctness of generated outputs~\citep{verifier}. Chain-of-thought (CoT) prompting has emerged as a particularly effective strategy, leveraging demonstrations of step-by-step reasoning to improve performance on complex tasks~\citep{wei2022chain,kojima2022large,reynolds2021prompt}. More recently, techniques such as intermediate self-reflection have been proposed to enable models to iteratively refine their reasoning processes~\citep{reflexion,madaan2023}. 

In parallel, supervised fine-tuning (SFT) has been employed to embed reasoning capabilities directly into LLMs. Studies such as~\citep{lewkowycz2022solving} and~\citep{rajani2019explain} demonstrate that fine-tuning on high-quality datasets can enhance problem-solving abilities. Notably, integrating CoT reasoning into SFT has shown significant promise; works like~\citep{zelikman2022star,s1simpletesttimescaling,ye2025limoreasoning} illustrate that fine-tuning with small, carefully curated datasets of CoT examples can yield substantial performance gains. However, these efforts have predominantly focused on large-scale LLMs, typically ranging from 7 billion to over 100 billion parameters. This reliance on massive models limits accessibility and practicality for resource-constrained settings, motivating the exploration of alternative approaches for smaller LLMs.

\subsection{Reasoning with Reinforcement Learning}
\label{subsec:rl_reasoning}
Reinforcement learning (RL) has emerged as a powerful paradigm for improving reasoning in LLMs, particularly for tackling complex, multi-step problems. Unlike SFT, which often optimizes for imitation of training data, RL enables models to learn from feedback, enhancing generalization to both in-distribution and out-of-distribution tasks~\citep{chu2025sftmemorizesrlgeneralizes,yeo2025demystifying}. Recent advancements underscore the efficacy of RL in this domain. For example,~\cite{o1} and~\cite{deepseekr12025} demonstrate that RL-based training can significantly boost reasoning performance, while~\cite{kimiteam2025kimik15scalingreinforcement} explores scaling laws for RL-driven LLMs. These studies highlight RL’s ability to refine decision-making processes by optimizing for task-specific rewards, such as correctness or logical coherence.

Despite these advances, RL-based methods are not without limitations. They typically demand substantial computational resources, often exceeding those required for SFT, and are predominantly applied to large LLMs. This focus on scale renders RL impractical for smaller models and restricts its adoption outside well-resourced organizations, such as major technology firms. Furthermore, privacy concerns arise when deploying such models, as self-hosting becomes infeasible for most academic or industrial entities with limited infrastructure. Consequently, there remains a critical gap in the literature: the application of RL to enhance reasoning in small LLMs under resource and privacy constraints.


\section{Limitations \& Discussion}
\label{sec:discussion}

While our study demonstrates the promise of RL-based fine-tuning for enhancing the reasoning abilities of small LLMs, several limitations and broader implications warrant discussion. These insights not only contextualize our findings but also highlight avenues for future research.

\subsection{Limitations}
First, our experiments were constrained by a 24-hour training window on a modest cluster of 4 NVIDIA A40 GPUs (48 GB VRAM each), limiting the number of global steps (e.g., 500 in Experiment 1 versus a potential 1500 for one epoch). This restriction curtailed our ability to fully explore the long-term behavior of the model, particularly beyond 200 steps, where performance degradation and multilingual outputs emerged. Second, the maximum completion length (4096 tokens in Experiment 1, reduced to 3584 in Experiments 2 and 3) proved insufficient for extremely hard problems in the \textbf{open-s1} dataset, forcing the model to truncate reasoning processes prematurely. This suggests that our methodology may underexploit the potential of small LLMs on complex tasks requiring extended reasoning chains.

Third, the multilingual nature of the base model, \texttt{DeepSeek-R1-Distill-Qwen-1.5B}, introduced unintended language drift after 150–200 steps, despite efforts to enforce English-only outputs via prompts in Experiment 3. This limitation reflects a trade-off in using a pre-trained, multilingual foundation, which, while efficient, complicates monolingual optimization. Finally, our evaluation focused exclusively on mathematical reasoning benchmarks, leaving the generalizability of our approach to other domains - such as scientific reasoning or coding - unexplored. These constraints highlight the need for cautious interpretation of our results within the specified scope.

\subsection{Discussion}
Our findings reveal a nuanced trade-off between efficiency and reasoning depth in small LLMs. The rapid performance gains observed in the first 50–100 steps across all experiments (Insight 1) suggest that small, high-quality datasets can effectively bootstrap reasoning capabilities, aligning with prior work on data efficiency in RL~\citep{chu2025sftmemorizesrlgeneralizes}. However, the subsequent degradation underscores a sensitivity to over-optimization under fixed length constraints, a challenge also noted in larger models like DeepSeek-R1~\citep{deepseekr12025}. Experiment 2’s success with mixed difficulty levels (Insight 2) indicates that curriculum-like strategies could mitigate this. Meanwhile, the cosine reward’s stabilizing effect in Experiment 3 (Insight 3) suggests a promising direction for controlling reasoning verbosity, though it sacrifices peak accuracy compared to Experiment 2.

Comparatively, our Open-RS variants achieved performance rivaling or exceeding state-of-the-art 1.5B models (e.g., \texttt{DeepScaleR-1.5B-Preview}) and even some 7B models, at a fraction of the cost and data volume. This efficiency challenges the prevailing reliance on massive datasets and computational resources in reasoning enhancement~\citep{o1,deepscaler2025}, offering a scalable alternative for resource-constrained environments. However, the persistent multilingual drift and length limitations point to inherent challenges in adapting multilingual base models and optimizing for complex tasks within tight bounds.

\subsection{Future Directions}
These limitations suggest several research avenues. First, extending training duration or employing multi-stage length schedules could address truncation issues, allowing the model to handle harder problems without compromising stability. Second, incorporating a lightweight language reward or monolingual pre-filtering of the base model might mitigate language drift, enhancing output consistency. Third, expanding the benchmark suite to include non-mathematical domains would test the generalizability of our approach, aligning with broader AGI goals. Finally, exploring hybrid methods - such as combining GRPO with search algorithms like MCTS~\citep{feng2024alphazeroliketreesearchguidelarge} - could further deepen reasoning capacity without significantly increasing resource demands.

In conclusion, our work demonstrates that RL-based fine-tuning can unlock substantial reasoning potential in small LLMs, even under stringent constraints. By identifying key trade-offs and offering practical insights, we pave the way for developing efficient, reasoning-capable models that balance performance and accessibility - a critical step toward democratizing advanced AI technologies.


\section{Datasets} \label{app:benchmarks}
Detail datasets are used in \cref{subsec:benchmarks}.
\begin{itemize}
\item \textbf{AIME24}~\footnote{\url{https://huggingface.co/datasets/AI-MO/aimo-validation-aime}}: 30 problems from the 2024 American Invitational Mathematics Examination, emphasizing advanced high-school-level reasoning.
\item \textbf{AMC23}~\footnote{\url{https://huggingface.co/datasets/AI-MO/aimo-validation-amc}}: 40 problems from the 2023 American Mathematics Competition, testing foundational mathematical skills.
\item \textbf{MATH-500}~\citep{lightman2023lets,hendrycksmath2021}: A subset of 500 problems from the MATH benchmark, sourced from various mathematics competitions and spanning algebra, calculus, and geometry.
\item \textbf{Minerva}~\citep{minervamath}: 272 undergraduate-level problems across physics, biology, chemistry, economics, and other sciences, requiring quantitative reasoning (mathematics subset used).
\item \textbf{OlympiadBench}~\citep{he-etal-2024-olympiadbench}: 675 Olympiad-level problems in mathematics and physics, designed to challenge advanced reasoning abilities.
\end{itemize}

Table~\ref{tab:datasets} summarizes the datasets and their sample sizes. This diverse collection ensures a comprehensive assessment of the model’s reasoning generalization across problem types and difficulty levels.

\begin{table}[ht]
\centering
\caption{Benchmark Datasets and Sample Sizes for Evaluation}
\label{tab:datasets}
\begin{tabular}{@{}l | r@{}}
\toprule
\textbf{Dataset} & \textbf{\# Samples} \\
\midrule
AIME24 & 30 \\
MATH-500 & 500 \\
AMC23 & 40 \\
Minerva & 272 \\
OlympiadBench & 675 \\
\bottomrule
\end{tabular}
\end{table}

\section{Baseline Models} \label{app:baseline-models}
The description of baseline models in \cref{subsec:baselines}.

\begin{itemize}
\item \textbf{General-Purpose Large Models}:
\begin{itemize}
   \item \texttt{Llama-3.1-70B-Instruct}~\citep{MetaAI2024}: A 70B-parameter model optimized for instruction-following.
   \item \texttt{o1-preview}~\citep{o1preview}: A high-performing reasoning model from OpenAI.
\end{itemize}
\item \textbf{Mathematics-Focused 7B Models}:
\begin{itemize}
   \item \texttt{Qwen-2.5-Math-7B-Instruct}~\citep{yang2024qwen25mathtechnicalreportmathematical}: An RL-trained model with a reward model for mathematical reasoning.
   \item \texttt{rStar-Math-7B}~\citep{guan2025rstar}: Uses Monte Carlo Tree Search (MCTS) for deep reasoning and self-evolution.
   \item \texttt{Eurus-2-7B-PRIME}~\citep{cui2025processreinforcementimplicitrewards}: Employs the PRIME method with online RL and process rewards.
   \item \texttt{Qwen2.5-7B-SimpleRL}~\citep{zeng2025simplerl}: Applies Proximal Policy Optimization (PPO) with rewards based on final answers.
\end{itemize}
\item \textbf{Mathematics-Focused 1.5B Models}:
\begin{itemize}
   \item \texttt{DeepSeek-R1-Distill-Qwen-1.5B}~\citep{deepseekr12025}: The original untrained baseline.
   \item \texttt{DeepScaleR-1.5B-Preview}~\citep{deepscaler2025}: Fine-tuned with GRPO on 40,000 math problem-answer pairs across multiple RL stages.
   \item \texttt{Still-3-1.5B-Preview}~\citep{Slow_Thinking_with_LLMs_3_Preview}: RL-trained with a focus on slow thinking (e.g., tree search) using 30,000 curated math examples.
\end{itemize}
\end{itemize}

\section{Hyperparameter Setup} \label{app:hyperparameters}
\cref{tab:hypers} show parameters that used in training phase.

\begin{table}[h]
\centering
\caption{Hyperparameter Setups for GRPO Trainer}
\begin{tabular}{ll}
\toprule
\textbf{Parameter} & \textbf{Value} \\
\midrule
\multicolumn{2}{l}{\textit{General Settings}} \\
bf16 & true \\
use\_vllm & true \\
vllm\_device & auto \\
vllm\_enforce\_eager & true \\
vllm\_gpu\_memory\_utilization & 0.7 \\
vllm\_max\_model\_len & 4608 \\
do\_eval & false \\
output\_dir & data/OpenRS-GRPO \\
overwrite\_output\_dir & true \\
\midrule
\multicolumn{2}{l}{\textit{Training Configuration}} \\
gradient\_accumulation\_steps & 4 \\
gradient\_checkpointing & true \\
gradient\_checkpointing\_kwargs & use\_reentrant: false \\
learning\_rate & 1.0e-06 \\
lr\_scheduler\_type & cosine\_with\_min\_lr \\
lr\_scheduler\_kwargs & min\_lr\_rate: 0.1 \\
warmup\_ratio & 0.1 \\
max\_steps & 500 \\
num\_train\_epochs & 1 \\
per\_device\_train\_batch\_size & 6 \\
per\_device\_eval\_batch\_size & 6 \\
\midrule
\multicolumn{2}{l}{\textit{Generation Settings}} \\
max\_prompt\_length & 512 \\
max\_completion\_length & 3584 or 4096 \\
num\_generations & 6 \\
temperature & 0.7 \\
seed & 42 \\
\midrule
\multicolumn{2}{l}{\textit{Logging and Saving}} \\
log\_completions & true \\
log\_level & info \\
logging\_first\_step & true \\
logging\_steps & 1 \\
logging\_strategy & steps \\
save\_strategy & steps \\
save\_steps & 50 \\
report\_to & wandb \\
\midrule
\multicolumn{2}{l}{\textit{Reward Configuration}} \\
reward\_funcs & format, accuracy (cosine) \\
reward\_weights & 1.0, 2.0 \\
\midrule
\multicolumn{2}{l}{\textit{Hub Settings}} \\
hub\_model\_id & OpenRS-GRPO \\
hub\_strategy & every\_save \\
push\_to\_hub & true \\
\bottomrule
\end{tabular}
\label{tab:hypers}
\end{table}

\end{document}